\documentclass[submission,copyright,creativecommons]{eptcs}
\usepackage{breakurl}             
\usepackage{underscore}           
\usepackage[dvipsnames]{xcolor}
\usepackage{amsthm}
\usepackage{amsfonts}
\usepackage[noend]{algorithm2e}
\usepackage{makecell}
\usepackage{multirow}
\usepackage{amsmath}
\usepackage{tikz}

\theoremstyle{definition}
\newtheorem{definition}{Definition}

\title{Extending the Service Composition Formalism \\ with Relational Parameters}
\author{Paul Diac, Liana \c Tuc\u ar, Radu Mereu\c t\u a
\institute{Alexandru Ioan Cuza Unviersity of Ia\c si, Rom\^ ania}
\email{\{paul.diac, liana.tucar, radu.mereuta\}@info.uaic.ro}
}

\begin{document}
\maketitle

\vspace{-0.5cm}
\begin{abstract}
Web Service Composition deals with the (re)use of Web Services to provide complex functionality, inexistent in any single service. Over the state-of-the-art, we introduce a new type of modeling, based on ontologies and relations between objects, which allows us to extend the expressiveness of problems that can be solved automatically.
\end{abstract}

\vspace{-0.5cm}
\section{Introduction}
Web Service Composition is a complex research area, involving other domains such as: web standards, service-oriented architectures, semantics, knowledge representation, algorithms, optimizations, and more \cite{rao2004survey}. We propose an extended model that allows the specification of relationships between parameters, as a generalization of previous models such as \cite{bansal2008wsc}. Moreover, it allows working with different instances of the same type of concept within the automatic composition; a feature that is fundamental in manual composition, and also defines \emph{inference rules}. The formalism defined in Section \ref{modelsection} is a complete specification of the model presented in our previous work \cite{diac2019}. We motivate the proposed model by an intuitive example and verify its effectiveness by implementing and testing a composition algorithm.

\vspace{-0.4cm}
\section{Motivation}

\vspace{-0.9cm}
\tikzset{every picture/.style={line width=0.75pt}} 
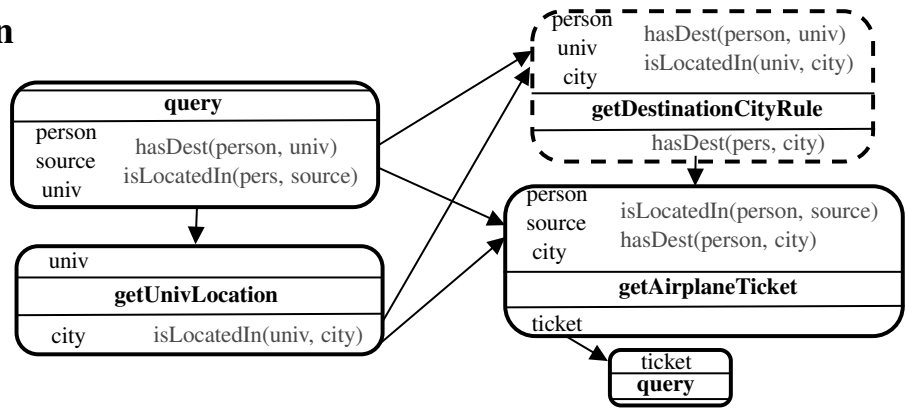
\hspace{3cm}
\begin{tikzpicture}[x=0.75pt,y=0.75pt,yscale=-1,xscale=1]

\draw  [line width=1.5]  (18.75,50.58) .. controls (18.75,43.63) and (24.38,38) .. (31.33,38) -- (190.78,38) .. controls (197.73,38) and (203.36,43.63) .. (203.36,50.58) -- (203.36,88.32) .. controls (203.36,95.27) and (197.73,100.9) .. (190.78,100.9) -- (31.33,100.9) .. controls (24.38,100.9) and (18.75,95.27) .. (18.75,88.32) -- cycle ;
\draw    (22,42) -- (200,42) ;

\draw    (18.75,57.71) -- (204.11,56.96) ;

\draw  [dash pattern={on 5.63pt off 4.5pt}][line width=1.5]  (281.14,17.18) .. controls (281.14,8.8) and (287.94,2) .. (296.32,2) -- (437.82,2) .. controls (446.2,2) and (453,8.8) .. (453,17.18) -- (453,62.72) .. controls (453,71.1) and (446.2,77.9) .. (437.82,77.9) -- (296.32,77.9) .. controls (287.94,77.9) and (281.14,71.1) .. (281.14,62.72) -- cycle ;
\draw    (280.77,43.45) -- (453.65,43.45) ;

\draw    (281.12,62.12) -- (454,61.37) ;

\draw  [line width=1.5]  (21,131.53) .. controls (21,125.51) and (25.88,120.62) .. (31.91,120.62) -- (194.69,120.62) .. controls (200.72,120.62) and (205.61,125.51) .. (205.61,131.53) -- (205.61,164.27) .. controls (205.61,170.3) and (200.72,175.19) .. (194.69,175.19) -- (31.91,175.19) .. controls (25.88,175.19) and (21,170.3) .. (21,164.27) -- cycle ;
\draw    (20.25,136.4) -- (205.61,136.4) ;

\draw    (20.62,155.44) -- (205.98,154.7) ;

\draw  [line width=1.5]  (267.46,105.5) .. controls (267.46,97.14) and (274.23,90.37) .. (282.58,90.37) -- (452.85,90.37) .. controls (461.2,90.37) and (467.97,97.14) .. (467.97,105.5) -- (467.97,150.87) .. controls (467.97,159.23) and (461.2,166) .. (452.85,166) -- (282.58,166) .. controls (274.23,166) and (267.46,159.23) .. (267.46,150.87) -- cycle ;
\draw    (266.64,133.56) -- (467.97,133.56) ;

\draw    (267.46,151.49) -- (468.79,150.74) ;

\draw    (111.99,100.9) -- (111.35,117.67) ;
\draw [shift={(111.27,119.67)}, rotate = 272.2] [fill={rgb, 255:red, 0; green, 0; blue, 0 }  ][line width=0.75]  [draw opacity=0] (8.93,-4.29) -- (0,0) -- (8.93,4.29) -- cycle    ;

\draw    (206,158) -- (279,31.73) ;
\draw [shift={(280,30)}, rotate = 480.03] [fill={rgb, 255:red, 0; green, 0; blue, 0 }  ][line width=0.75]  [draw opacity=0] (8.93,-4.29) -- (0,0) -- (8.93,4.29) -- cycle    ;

\draw    (204,69.33) -- (279.3,23.05) ;
\draw [shift={(281,22)}, rotate = 508.42] [fill={rgb, 255:red, 0; green, 0; blue, 0 }  ][line width=0.75]  [draw opacity=0] (8.93,-4.29) -- (0,0) -- (8.93,4.29) -- cycle    ;

\draw    (363.5,75) -- (363.3,88.33) ;
\draw [shift={(363.27,90.33)}, rotate = 270.86] [fill={rgb, 255:red, 0; green, 0; blue, 0 }  ][line width=0.75]  [draw opacity=0] (8.93,-4.29) -- (0,0) -- (8.93,4.29) -- cycle    ;

\draw    (204,170) -- (265.98,117.3) ;
\draw [shift={(267.5,116)}, rotate = 499.62] [fill={rgb, 255:red, 0; green, 0; blue, 0 }  ][line width=0.75]  [draw opacity=0] (8.93,-4.29) -- (0,0) -- (8.93,4.29) -- cycle    ;

\draw  [line width=1.5]  (320.25,178.6) .. controls (320.25,175.51) and (322.75,173) .. (325.85,173) -- (375.65,173) .. controls (378.75,173) and (381.25,175.51) .. (381.25,178.6) -- (381.25,195.4) .. controls (381.25,198.49) and (378.75,201) .. (375.65,201) -- (325.85,201) .. controls (322.75,201) and (320.25,198.49) .. (320.25,195.4) -- cycle ;
\draw    (320,184.78) -- (381.25,184.78) ;

\draw    (298,166) -- (317.53,177.58) ;
\draw [shift={(319.25,178.6)}, rotate = 210.67000000000002] [fill={rgb, 255:red, 0; green, 0; blue, 0 }  ][line width=0.75]  [draw opacity=0] (8.93,-4.29) -- (0,0) -- (8.93,4.29) -- cycle    ;

\draw    (203.5,81) -- (265.68,109.17) ;
\draw [shift={(267.5,110)}, rotate = 204.38] [fill={rgb, 255:red, 0; green, 0; blue, 0 }  ][line width=0.75]  [draw opacity=0] (8.93,-4.29) -- (0,0) -- (8.93,4.29) -- cycle    ;

\draw    (320,197.78) -- (381.25,197.78) ;

\draw (110.13,50.47) node [scale=0.8] [align=left] {\textbf{query}};
\draw (46.37,78.68) node [scale=0.8] [align=left] {person\\source\\ \ univ};
\draw (369.51,52.19) node [scale=0.8] [align=left] {\textbf{getDestinationCityRule}};
\draw (386.71,69.41) node [scale=0.8,color={rgb, 255:red, 74; green, 74; blue, 74 }  ,opacity=1 ] [align=left] { hasDest(pers, city) };
\draw (306.46,22.72) node [scale=0.8] [align=left] {person\\ \ univ\\ \ \ city};
\draw (111.88,146.07) node [scale=0.8] [align=left] {\textbf{getUnivLocation}};
\draw (143.07,166.04) node [scale=0.8,color={rgb, 255:red, 74; green, 74; blue, 74 }  ,opacity=1 ] [align=left] {isLocatedIn(univ, city)};
\draw (47.46,127.85) node [scale=0.8] [align=left] {univ};
\draw (370.02,142.05) node [scale=0.8] [align=left] {\textbf{getAirplaneTicket}};
\draw (293.85,159.53) node [scale=0.8] [align=left] {ticket};
\draw (293.8,111.09) node [scale=0.8] [align=left] {person\\source\\ \ city};
\draw (47.46,166.85) node [scale=0.8] [align=left] {city};
\draw (348.46,192) node [scale=0.8] [align=left] {\textbf{query}};
\draw (348.9,178.85) node [scale=0.8] [align=left] {ticket};
\draw (134.37,78.68) node [scale=0.8] [align=left] { \ \ \textcolor[rgb]{0.29,0.29,0.29}{hasDest(person, univ) }\\\textcolor[rgb]{0.29,0.29,0.29}{isLocatedIn(pers, source)}};
\draw (390.46,21.72) node [scale=0.8,color={rgb, 255:red, 74; green, 74; blue, 74 }  ,opacity=1 ] [align=left] {hasDest(person, univ)\\isLocatedIn(univ, city)};
\draw (390.8,111.09) node [scale=0.8,color={rgb, 255:red, 74; green, 74; blue, 74 }  ,opacity=1 ] [align=left] {isLocatedIn(person, source)\\hasDest(person, city)};
\end{tikzpicture}

We present a simple query, where a user wants to travel to a university located in a different city. Each rectangle represents a web service with input at the top and output at the bottom. We also represent the query twice as a services with no input and respectively, no output. The dashed rectangle is an inference rule, handled by the algorithm as a virtual web service.

Because we cannot directly get the answer to the query, we must use the information provided by different services and rules found in the ontology to compose an answer. We see that in order to buy a ticket we must know the source city and the destination city, the latter found indirectly. We must first call a different web service which finds the city where the university is located and by using the inference rule we can finally match the precondition of the web service which can return the plane ticket. Arrows show parameters and relations matching. The algorithm in \ref{algorithm} finds the correct order of calls; in this case: \emph{(1) getUniversityLocation}, applies \emph{(2) getDestinationCityRule} and then \emph{(3) getAriplaneTicket}.

\section{Formal Model}\label{modelsection}
\vspace{-0.1cm}
We define the model in three steps: the original composition formalism matches parameters by name; the semantic level defining concepts over a taxonomy; and finally the new relational level, enhancing the taxonomy to a full ontology. On these three levels, expressiveness increases allowing for more and more natural composition examples to be resolvable by composition algorithms if appropriately modeled. The first two are well-known and were used in the Composition Challenges in \cite{blake2005eee} and \cite{bansal2008wsc}. The last level is our contribution, introducing two important concepts: parameter relations and, as a consequence of the first, type instances as separate matching objects.

\vspace{-0.2cm}
\subsection{Name-based Parameter Matching}
\vspace{-0.1cm}
The initial and simplest model for \emph{Web Service Composition} uses parameter \textbf{names} to match services. Each \textbf{name} represents a \textbf{concept}. Expecting that parameters are chosen from a set of predefined concepts, the output of a previously called service can be used as input to another service. The user specifies a composition request by a list of known concepts and a list of required concepts. A request has the structure of a service but expresses the need for such a service. A satisfying composition is a (partially ordered) list of services generating the requested concepts starting from known concepts.

\begin{definition}{\textbf{Concept}s}\label{conceptdef}
are elements from the predefined set of all concepts $\mathbb{C}$.
\end{definition}

\begin{definition}{\textbf{Web Service}s}\label{servicedef}
 are triplets $\langle name, I, O \rangle$ consisting of: the service name and two disjoint sets of concepts, also referred to as parameters: input and output. The \textbf{User Request} has the same structure and specifies a required functionality, possibly solvable by a list of services. $I \cap O = \emptyset$ and $I,O \subseteq \mathbb{C}$.
\end{definition}

\begin{definition}{The \textbf{Repository}} is the set of all services, also written as $\mathbb{S}$.
\end{definition}

\begin{definition}{\textbf{Parameter Matching}.}\label{paramdef}
If $C$ is a set of (known) concepts and $ws = \langle ws.name, ws.I, ws.O \rangle$ is a web service, then $C$ matches $ws$ iff $ws.I \subseteq C$.
\\ The result of matching, or the union of $C$ and $ws.O$ is $C \oplus ws = C \cup ws.O$.
\end{definition}

\begin{definition}{\textbf{Chained Matching}.}
If $C$ is a set of concepts and $\langle ws_{1}, ws_{2}, \dots, ws_{n} \rangle$ a list of services then: $C \oplus ws_{1} \oplus ws_{2} \oplus \dots \oplus ws_{n} $ is a chain of matching services, generating $C \cup \bigcup_{i=1}^{n} ws_{i}.O$; valid iff:
$$ws_i.I \subseteq \Bigg( C \cup \bigg(\bigcup_{j=1}^{i-1} ws_{j}.O\bigg) \Bigg), \forall i = \overline{1..n}$$
\end{definition}

\begin{definition}{\textbf{Web Service Composition Problem}}. Given a repository of services $\mathbb{S}$  and a user request $r = \langle r.name, r.I, r.O\rangle$, all with parameters defined over the set of concepts $\mathbb{C}$; find a chain of matching services $\langle ws_{1}, ws_{2}, \dots, ws_{n} \rangle$ such that $r.I \oplus ws_{1} \oplus ws_{2} \oplus \dots \oplus ws_{n} \subseteq r.O$.
\end{definition}

\vspace{-0.2cm}
\subsection{Taxonomy-based Parameter Matching}\label{taxonomymodel}
\vspace{-0.1cm}
Subsequent models extend the definitions of \textbf{concept}s and \textbf{parameter matching} (\ref{conceptdef} and \ref{paramdef}), and the rest of the definitions adapt to these changes.

\begin{definition}{\textbf{Concept}s (in model \ref{taxonomymodel})}
are elements of the set of concepts $\mathbb{C}$, over which the binary relation $subtypeOf$ is defined. $subtypeOf \subseteq \mathbb{C}^{\hspace{1px}2}$ and $subtypeOf$ is \emph{transitive}.
\end{definition}

\begin{definition}{\textbf{Parameter Matching} (in \ref{taxonomymodel}).}
If $C \in \mathbb{C}$ is a set of concepts with $subtypeOf$ relation, and $ws = \langle ws.name, ws.I, ws.O \rangle$ a service, then $C$ matches $ws$ iff: \\ $\forall \hspace{3px} c \in ws.I, \hspace{5px} \exists \hspace{5px} spec \in C, \hspace{5px} such \hspace{5px} that \hspace {5px} (spec, c) \in subtypeOf$.

\noindent
The result of matching is $C \oplus ws = C \hspace{3px} \bigcup \hspace{3px} \big\{gen \in \mathbb{C} \hspace{3px} \big| \hspace{3px} \exists \hspace{3px} c \in ws.O, \hspace{5px} such \hspace{5px} that \hspace{5px} (c, gen) \in subtypeOf \big\}$.
\end{definition}

\vspace{-0.2cm}
\subsection{Ontological Level: Relational and Contextual
Model}\label{ontologymodel}
\vspace{-0.1cm}

The main contribution of the paper is the introduction of two elements: \emph{relations} and \emph{objects}.
Relations are a generalization of the $subtypeOf$ relation of the previous level. Multiple relations are allowed between concepts, defined in the semantic \emph{ontology}. Service providers do not define new relations; they can only use existing relations defined in the \emph{ontology} to describe their parameters. Relations can be \emph{transitive} and/or \emph{symmetric}.

Concepts can now be described with more semantic context, and it is useful to allow updates on it. Therefore, we also introduce \emph{objects}, that are similar to instances of concepts. Instances are not concrete values of concept types, but distinct elements that are passed trough service workflow, distinguished by their provenance and described by a set of semantic relations.

\emph{Inference rules} are also introduced as a generalization of relation properties. \emph{Inference rules} generate new relations on objects if some preconditions are met. Similarly, web service calls that exclusively generate objects can also generate relations. Service input can define preconditions that include relations on objects matching input parameters.

\begin{definition}{An \textbf{Object} is}
an element of the set of objects $\mathbb{O} = \{ o = \langle id, type \rangle \}$. $id$ is a unique identifier generated at object creation. The $type$ is a concept: $type \in \mathbb{C}$.
\end{definition}

\begin{definition}{\textbf{Relation}.}\label{relationdef}
A relation $r$ is a triple consisting of: the name as a unique identifier, relation properties, and the set of pairs of objects that are in that relation. The latter is dynamic, i.e., can be extended through the composition process.
\vspace{-5px}
$$ \mathbb{R} = \big\{\langle \hspace{1px} name, properties, objects \hspace{1px} \rangle \hspace{2px} \big| \hspace{2px} properties \subseteq \{ transitivity, \hspace{2px} symmetry\} \hspace{3px} and \hspace{3px} objects \subseteq \mathbb{O}^{\hspace{1px}2}\big\} $$
\end{definition}

\begin{definition}{The \textbf{knowledge} $\mathbb{K}$}
 is a dynamic structure consisting of objects and relations between the objects. Knowledge describes what is known at a stage of the composition workflow, i.e., at a time when a set of services have been added to the composition. $\mathbb{K} = \langle \hspace{1px} \mathbb{O}, \mathbb{R} \hspace{1px}\rangle$.
\end{definition}

\begin{definition}{\textbf{Web Services} (in model \ref{ontologymodel}) are tuples $\langle name, I, O, relations \rangle$ with $I, O$ defined as in def. \ref{servicedef} and $relations$ specifying preconditions and postconditions (effects) over objects matched to service inputs or generated at output. $relations$ within service definitions are pairs consisting of: the $name$ used to refer to an existing relation (the relation from $\mathbb{R}$ with the same name), and a binary relation over all service parameters. Relations between inputs are preconditions, and relations between output are effects, i.e., they are generated after the call. Relations between input and output parameters are effects.}
$$ws.relations = \big\{\langle \hspace{1px} name, parameters \hspace{1px} \rangle \hspace{2px} \big| \hspace{2px} names \hspace{3px} \mathit{from} \hspace{3px} \mathbb{R} \hspace{3px} and \hspace{3px} parameters \subseteq (ws.I \cup ws.O)^{2} \hspace{1px} \big\}$$
\end{definition}

\begin{definition}{\textbf{Inference Rules} (in \ref{ontologymodel}) are tuples $rule = \langle \hspace{0.5px} name, parameters, preconditions, \mathit{effects} \rangle$ where $parameters$ is a set of parameter names with local visibility (within rule), and preconditions and effects are relations defined over $parameters$. More precisely: \\ $$rule.preconditions, rule.\mathit{effects} \subseteq \big\langle \bigcup\limits_{rel\hspace{1px}\in\hspace{1px}\mathbb{R}} rel.name, rule.parameters^{\hspace{1px}2} \big\rangle$$
\\The set of all inference rules is written as $\mathbb{I}$. Preconditions must hold before applying the rule for the objects matching rule parameters, and relations in the effects are generated accordingly. Rules are structurally similar to services, but they apply automatically and, conceptually, with no cost. For example, \emph{transitivity} and \emph{symmetry} are particular rules, the following expresses that $equals$ is symetric:}
$$equals_{symmetric} = \big\langle \{X, Y\}, \{\langle equals, \{ (X, Y)\} \rangle\},  \{\langle equals, \{ (Y, X)\} \rangle\}  \big\rangle$$
\end{definition}

\begin{definition}{\textbf{Ontology} $\mathbb{G}$ consists of: concepts organized hierarchically, relations and inference rules.}
$\mathbb{G} = \langle \hspace{3px} \mathbb{C}, subtypeOf, \mathbb{R}, \mathbb{I} \hspace{3px} \rangle$. At ontological level, relations are static and defined only by names and properties. At \textbf{knowledge} level, relations are dynamic in what objects they materialize to. We refer to both using $\mathbb{R}$.
\end{definition}

\begin{definition}{\textbf{Parameter Matching} (in \ref{ontologymodel}). In ontology $\mathbb{G} = \langle \hspace{3px} \mathbb{C}, subtypeOf, \mathbb{R}, \mathbb{I} \hspace{3px} \rangle$, a web service $ws$ matches (is "callable" in) a knowledge state $\mathbb{K} = \langle \hspace{1px} \mathbb{O}, \mathbb{R} \hspace{1px}\rangle$, iff:}

$$\exists \hspace{3px} function \hspace{5px} f : ws.I \rightarrow \mathbb{O} \hspace{4px} such \hspace{3px} that:$$

\vspace{-23px}

\begin{equation}
\begin{split}
\forall \hspace{3px} i \in ws.I, \big(f(i),i\big) \in subtypeOf \hspace{5px} and
\end{split}
\end{equation}

\vspace{-23px}

\begin{equation}
\begin{split}
\forall \hspace{3px} \big( \hspace{3px} i, j \in ws.I \hspace{5px} and \hspace{5px} r_{ws} \in ws.relations, \hspace{3px} with \hspace{5px} (i,j) \in r_{ws}.parameters \big) \\
\exists \hspace{3px} \big( r_{obj} \in \mathbb{R} \hspace{5px} with: \hspace{5px} r_{obj}.name=r_{ws}.name \hspace{5px} and \hspace{5px} \big(f(i),f(j)\big) \in r_{obj}.objects\big)
\end{split}
\end{equation}
\end{definition}

We skip other similar definitions in model \ref{ontologymodel} that are intuitively similar, such as $\mathbb{K} \oplus ws$, \textbf{chained matching}, \textbf{user request} and the \textbf{composition problem}.

\vspace{-4px}
\section{Composition Algorithm}
\label{algorithm}
\vspace{-2px}
\paragraph{Overview}
The algorithm takes as input a query from the user, the repository and ontology and returns a composition of services that answers the query. We start by populating a set (called "knowledge") with objects and relations based on the information provided by the user. We then repeat the process of adding new objects and relations until no more service calls can be made or until the query can be answered. 
	
	
	\begin{algorithm}[H]
		$init$: data structures and create virtual services for inference rules\;
		\While{$\neg canAnswerQuery(query)$ And $\mathit{compositionUpdated} = True$}{
			$\mathit{compositionUpdated} \gets False$ \;
			\ForEach{$service \in repository$}{
				$possibleCalls \gets searchForPossibleCalls(service)$ \;
				\ForEach{$servCall \in possibleCalls$}{
					\If{$\mathit{providesUsefulInformation}(servCall)$}{
						$makeCall(servCall);$ $\mathit{compositionUpdated} \gets True$ \;	
					}
				}
			}
		}
		
		\lIf{$canAnswerQuery(query)$}{ \Return $composition$ 
		}\lElse{	\Return $Not\ Solved$ }
		\vskip -0.5cm
	\end{algorithm}

	\paragraph{Construction Phase}
	At each step, we iterate over all the services and search for all possible calls. We then add to the composition the service calls that provide new information: i.e. a service call is excluded if all the new objects added are semantically similar to others already present in the knowledge. To obtain the similarity between objects, we represent the knowledge as a labeled directed graph where vertices are objects (labeled with the type of the object), and edges are relations, and we consider two objects similar if their associated connected components are isomorphic.
	
	When a service call is made, new objects and relations corresponding with the service output and postconditions are created and added in the knowledge.
	
	\vspace{-0.2cm}
	\paragraph{Search for service calls} \label{search serv calls}
	Finding all possible service calls of a given service $serv$ means finding all combinations of objects that can be used as input parameters for the service: i.e. finding for each input parameter a corresponding object in the $knowledge$ that has a type that is equal or more general with the parameter type. Besides this condition regarding the types, all relations from preconditions need to hold between corresponding objects used to call the service.
	
	This problem reduces to finding all subgraph isomorphisms in the following problem instance: 
	\begin{itemize}
		\item $q = (V, E, L)$, where $V = serv.inputParams, E = serv.preConditions, \\
		L(u) = u.type, \forall u \in V$ and $L(e) = e.name, \forall e \in E$;
		\item $g = (V', E', L')$, where $V' = knwoledge.objects, E' = knowledge.relationsBo, \\
		L(u') = \{type\ \vert\ (u'.type, type) \in ontology.subType\}, \forall u' \in V'$ \\
		and $L(e') = e'.name, \forall e' \in E'$;
	\end{itemize}
	The associated decision problem is known to be NP-Complete and an optimized backtracking procedure was implemented to solve it. In real-world use cases, we expect that instances that are computationally hard are rare. This is because service and rule preconditions are checked at each step of the backtracking, pruning many execution paths. Moreover, the \emph{inference rules} that are more generic - i.e. without typed parameters - are defined in the ontology that cannot be updated by service developers or other users, so expensive rules can be safely avoided.
	
    \vspace{-0.1cm}
	\paragraph{canCallService}	
	To check if service calls provide useful information and if the query is solved, virtual services with corresponding parameters and conditions are constructed and checked if can be called. This problem is similar to the one described above, except only one solution is needed, not all possible service calls.
    
    An optimization implemented in the backtracking procedure is to \textbf{split the query graph into connected components} and to search each of them independently in the data graph. If the query graph is formed by multiple connected components this optimization helps by reducing a cartesian product of tested solutions of each component to a union of them. This has been proven to have a significant benefit on the runtime on tests (depending on how the tests are generated).

\vspace{-0.2cm}
\section{Conclusion}
\vspace{-0.1cm}

Current Web Service Composition models include limited semantics in expressing how service parameters are matched. Particularly, there is no way to express any relationships between parameters, and parameter typing models do not allow distinguishing the separation between instances of the same concept. In this paper, we propose a formalism that solves both of these limitations. We also implemented an efficient automatic composition algorithm that produced valid compositions on generated tests, using all elements in the proposed model.
\vspace{-0.2cm}

\nocite{*}
\bibliographystyle{eptcs}
\bibliography{FROMbibleo}
\end{document}